\documentclass[11pt,letterpaper]{article}
\usepackage{emnlp2017}
\usepackage{times}
\usepackage{latexsym}
\usepackage{url}
\usepackage{graphicx}
\usepackage{mathtools}
\usepackage{pgfplots}
\usepgfplotslibrary{external}
\usepackage{booktabs}
\usepackage{multirow}
\usepackage{xspace}
\usepackage{adjustbox}

\emnlpfinalcopy

\def\emnlppaperid{***}

\newcommand\conll{CoNLL'12\xspace}

\graphicspath{{fig/}}

\title{Revisiting Selectional Preferences for Coreference Resolution}

\author{
Benjamin Heinzerling\thanks{\hspace{0.7em}These authors contributed equally to this work.} \\
AIPHES \\
Heidelberg Institute for \\
Theoretical Studies \\
{\tt \small benjamin.heinzerling@h-its.org}
\And
Nafise Sadat Moosavi\footnotemark[1] \\
Heidelberg Institute for \\
Theoretical Studies \\
{\tt \small nafise.moosavi@h-its.org}
\And
Michael Strube \\
Heidelberg Institute for \\
Theoretical Studies \\
{\tt \small michael.strube@h-its.org}
}

\date{}

\begin{document}

\maketitle

\begin{abstract}
Selectional preferences have long been claimed to be essential for coreference resolution.
However, they are mainly modeled only implicitly by current coreference resolvers.
We propose a dependency-based embedding model of selectional preferences which allows fine-grained compatibility judgments with high coverage.
We show that the incorporation of our model improves coreference resolution performance on the CoNLL dataset,
matching the state-of-the-art results of a more complex system. 
However, it comes with a cost that makes it debatable
how worthwhile such improvements are.
\end{abstract}

\section{Introduction}
\label{sec:intro}

Selectional preferences have long been claimed to be useful for coreference resolution. In his seminal work on ``Resolving Pronominal References'' \newcite{hobbs78} proposed a semantic approach that requires reasoning about the ``demands the predicate makes on its arguments.''
For example, selectional preferences allow resolving the pronoun \emph{it} in the text \emph{``The Titanic hit an iceberg. It sank quickly.''}
Here, the predicate \emph{sink} `prefers' certain subject arguments over others: It is plausible that a ship sinks, but implausible that an iceberg does.

Work on the automatic acquisition of selectional preferences has shown considerable progress \cite{dagan90,resnik93,agirre01,pantel07,erk07,ritter10,vandecruys14}.
However, today's coreference resolvers \cite[i.a.]{martschat15c,wiseman16,clarkkevin16b}
capture selectional preferences only implicitly at best, e.g., via a given mention's dependency governor and other contextual features.

Since negative results do not often get reported, there is no clear evidence in the literature regarding the non-utility of particular knowledge sources. 
Consequently, an absence of the explicit modeling of selectional preferences in the recent literature is an indicator that incorporating this knowledge source has not been very successful for coreference resolution. 

More than ten years ago, \newcite{kehler04a} declared the ``non-utility of predicate-argument structures for pronoun resolution'' and observed that minor improvements on a small dataset were due to fortuity rather than selectional preferences having captured meaningful world knowledge relations. 
%\newcite{durrett13b} called integrating semantics into coreference resolution an ``uphill battle'' and  \newcite{strube15} reports that any attempt at incorporating world knowledge into a state-of-the-art system \cite{martschat15c} degraded performance.

\begin{figure*}[t!]
    \centering
	\includegraphics[width=\textwidth,height=4.3cm]{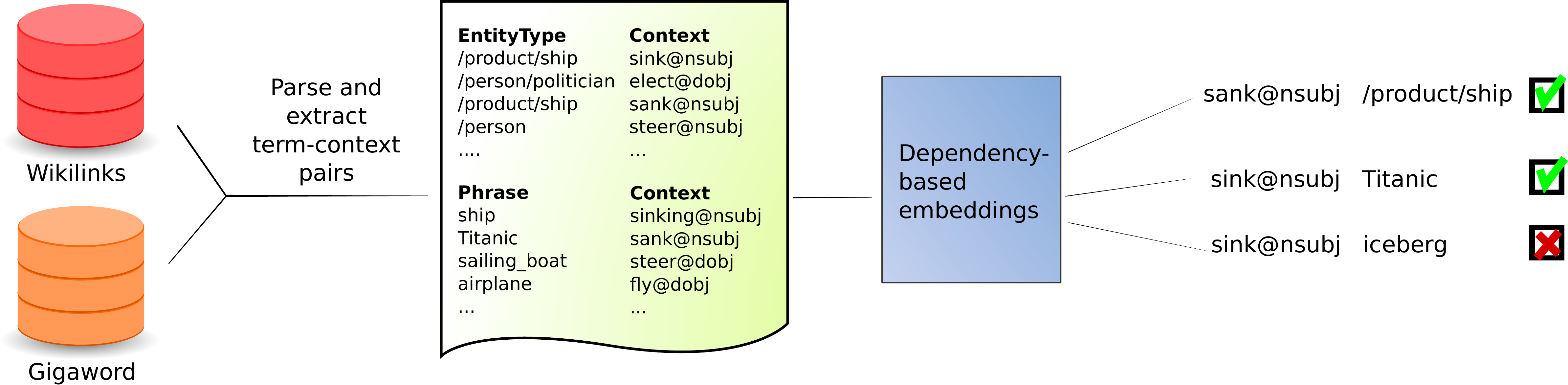}
	\caption{Dependency-based embedding model of selectional preferences.}
	\label{fig:overview}
\end{figure*}

The claim by \newcite{kehler04a} is based on selectional preferences extracted from a, by current standards, small number of $2.8$m predicate-argument pairs. 
Furthermore, they employ a simple (linear) maximum entropy classifier, which requires manual definition of feature combinations and is unlikely to fully capture the complex interaction between selectional preferences and other coreference features. 
%\newcite{durrett13b} achieve high coverage by extracting selectional preferences from a much larger corpus, but use a coarse model comprising only 20 latent clusters, such as "things which announce".
Therefore, it is worth revisiting how a better selectional preference model affects the performance of a more complex coreference resolver.

In this work, we propose a fine-grained, high-coverage model of selectional preferences and study its impact on a state-of-the-art, non-linear coreference resolver.
We show that the incorporation of our selectional preference model improves the performance.
%However, considering the additional resources that are required for training the selectional preference model,
However, it is debatable whether such small improvements, that cost notable extra time or resources, are advantageous.

% fine-grained entity types, and a large vocabulary into a single embedding space. 
% Therefore it has a high coverage. 
% (2) Selectional preferences and the coreference relation do not have a linear relation. 
% The use of hidden layers in our baseline coreference resolver, i.e.\ deep-coref by \newcite{clarkkevin16a}, 
% allows a non-linear combination of selectional preferences and other determining knowledge sources to make the final decision.

% 

\section{Modeling Selectional Preferences}
\label{sec:model}

The main design choice when modeling selectional preferences is the selection of a relation inventory, i.e.\ the concepts and entities that can be relation arguments, and the semantic relationships that hold between them. 

Prior work has studied many relation inventories. Predicate-argument statistics for word-word pairs (\emph{eat, food})\footnote{Examples due to \newcite{agirre01}.}
are easy to obtain but do not generalize to unseen pairs \cite{dagan90}. Class-based approaches generalize via word-class pairs (\emph{eat, /nutrient/food}) \cite{resnik93}  
or class-class pairs (\emph{/ingest, /nutrient/food}) \cite{agirre01}, 
but require disambiguation of words to classes and are limited by the coverage of the lexical resource providing such classes (e.g.\ WordNet).

Other possible relation inventories include semantic representations such as FrameNet frames and roles, event types and arguments, or abstract meaning representations. 
While these semantic representations are arguably well-suited to model meaningful world knowledge relationships, 
automatic annotation is limited in speed and accuracy, making it difficult to obtain a large number of such ``more semantic'' predicate-argument pairs. 
In comparison, syntactic parsing is both fast and accurate, making it trivial to obtain a large number of accurate, albeit ``less semantic'' predicate-argument pairs.
The drawback of a syntactic model of selectional preferences is susceptibility to lexical and syntactic variation. For example, \emph{The Titanic sank} and \emph{The ship went under} differ lexically and syntactically, but would have the same or a very similar representation in a semantic framework such as FrameNet.

Our model of selectional preferences (Figure~\ref{fig:overview}) overcomes this drawback via distributed representation of predicate-argument pairs, using (syntactic) dependencies that were specifically designed for semantic downstream tasks, and by resolving named entities to their fine-grained entity types.

\textbf{Distributed representation.} Inspired by Structured Vector Space \cite{erk08}, we embed predicates and arguments into a low-dimensional space in which (representations of) predicate slots are close to (representations of) their plausible arguments, as should be arguments that tend to fill the same slots of similar predicates, and predicate slots that have similar arguments. For example, \emph{captain} should be close to \emph{pilot}, \emph{ship} to \emph{airplane}, the subject of \emph{steer} close to both \emph{captain} and \emph{pilot}, and also to, e.g., the subject of \emph{drive}. Such a space allows judging the plausibility of unseen predicate-argument pairs.\footnote{%
Prior work generalizes to unseen predicate-argument pairs via WordNet synsets \cite{resnik93}, a generalization corpus \cite{erk07}, or tensor factorization \cite{vandecruys10}. Closest to our approach is neural model by \newcite{vandecruys14}, which, however, has much lower coverage since it is limited to 7k verbs and 30k arguments.}

We construct this space via dependency-based word embeddings \cite{levy14}. To see why this choice is better-suited for modeling selectional preferences than alternatives such as word2vec \cite{mikolov13b} or GloVe \cite{pennington14}, consider the following example:

\begin{center}
	\small
	\begin{tabular}{ccccc}
		captain & $\xleftarrow{\text{nsubj}}$ & steers & $\xrightarrow{\text{dobj}}$ & ship \\
		:: & & & & :: \\
		pilot & $\xleftarrow{\text{nsubj}}$ & steers & $\xrightarrow{\text{dobj}}$ & airplane \\
	\end{tabular}
\end{center}

\noindent Here, \emph{captain} and \emph{ship}, have high syntagmatic similarity, i.e., these words are semantically related and tend to occur close to each other. This also holds for \emph{pilot} and \emph{airplane}. In contrast, \emph{captain} and \emph{pilot}, as well as \emph{ship} and \emph{airplane} have high paradigmatic similarity, i.e., they are semantically similar and occur in similar contexts.
A model of selectional preferences requires paradigmatic similarity: The representations of \emph{captain} and \emph{pilot} in such a model should be similar, since they both can plausibly fill the subject slot of the predicate \emph{steer}. Due to their use of linear context windows, word2vec and GloVe capture syntagmatic similarity, while dependency-based embeddings capture paradigmatic similarity \cite[cf.][]{levy14}.

\textbf{Enhanced++ dependencies.} Due to distributed representation, our model generalizes over syntactic variation such as active/passive alternations: For example, \emph{steer@dobj}\footnote{%
In this work, a predicate's argument slots are denoted \emph{predicate@slot}.} 
is highly similar to \emph{steer@nsubjpass} (see Appendix for more examples). To further mitigate the effect of employing syntax as a proxy for semantics, we use Enhanced++ dependencies \cite{schuster16}. Enhanced++ dependencies aim to support semantic applications by modifying syntactic parse trees to better reflect relations between content words. For example, the plain syntactic parse of the sentence \emph{Both of the girls laughed} identifies \emph{Both} as subject of \emph{laughed}. The Enhanced++ representation introduces a subject relation between \emph{girls} and \emph{laughed}, which allows learning more meaningful selectional preferences: Our model should learn that girls (and other humans) laugh, while learning that an unspecified \emph{both} laughs is not helpful.

\textbf{Fine-grained entity types.} A good model of selectional preferences needs to generalize over named entities. For example, having encountered sentences like \emph{The Titanic sank}, our model should be able to judge the plausibility of an unseen sentence like \emph{The RMS Lusitania sank}.
For popular named entities, we can expect the learned representations of \emph{Titanic} and \emph{RMS Lusitania} to be similar, allowing our model to generalize, i.e., it can judge the plausibility of \emph{The RMS Lusitania sank} by virtue of the similarity between \emph{Titanic} and \emph{RMS Lusitania}. However, this will not work for rare or emerging named entities, for which no, or only low-quality, distributed representations have been learned.
To address this issue, we incorporate fine-grained entity typing \cite{ling12}.
For each named entity encountered during training, we generate an additional training instance by replacing the named entity with its entity type, e.g. \emph{(Titanic, sank@nsubj)} yields \emph{(/product/ship, sank@nsubj)}.

\section{Implementation}

We train our model by combining term-context pairs from two sources.
Noun phrases and their dependency context are extracted from GigaWord \cite{gigaword5.0data} and entity types in context from Wikilinks \cite{singh12}. Term-context pairs are obtained by parsing each corpus with the Stanford CoreNLP dependency parser \cite{manning14}. After filtering, this yields ca.\,1.4 billion phrase-context pairs such as \emph{(Titanic, sank@nsubj)} from GigaWord and ca.\,12.9 million entity type-context pairs such as \emph{(/product/ship, sank@nsubj)} from Wikilinks.
Finally, we train dependency-based embeddings using the generalized word2vec version by \newcite{levy14}, obtaining distributed representations of selectional preferences.
To identify fine-grained types of named entities at test time, we first perform entity linking using the system by \newcite{heinzerling16}, then query Freebase \cite{bollacker08} for entity types and apply the mapping to fine-grained types by \newcite{ling12}.

The plausibility of an argument filling a particular predicate slot can now be computed via the cosine similarity of their associated embeddings.
For example, in our trained model, the similarity of \emph{(Titanic, sank@nsubj)} is 0.11 while the similarity of \emph{(iceberg, sank@nsubj)} is -0.005, indicating that an iceberg sinking is less plausible.

\section{Do Selectional Preferences Benefit Coreference Resolution?}
\label{sec:coref-interaction}
\begin{table*}[t]
    \centering
    \resizebox{\textwidth}{!}{
        \begin{tabular}{l|rrr|rrr|rrr|r|rrr}
        \toprule
        \multicolumn{1}{c}{} & \multicolumn{3}{c}{MUC} & \multicolumn{3}{c}{$B^3$} & \multicolumn{3}{c}{CEAF$_e$} & CoNLL & \multicolumn{3}{c}{LEA} \\ \hline
        \multicolumn{1}{c|}{}&R & P & F$_1$ & R & P & F$_1$ & R & P & F$_1$ & Avg. F$_1$ & R & P & F$_1$\\
        \midrule
         baseline & $70.09$ & $80.01$ & $74.72$ & $57.64$ & $70.09$ & $63.26$ & $54.47$ & $63.92$ & $58.82$ & $65.60$ & $54.02$ & $66.45$ & $59.59$\\ \hline
         $-$gov & $70.10$ & $79.96$ & $74.71$ & $57.51$ & $70.31$ & $63.27$ & $54.41$ & $64.08$ & $58.85$ & $65.61$ & $53.93$ & $66.76$ & $59.66$  \\ 
         +SP & $70.85$ & $79.31$ & $74.85$ & $58.93$ & $69.16$ & ${63.64}$ & $55.25$ & $63.78$ & ${59.21}$ & ${65.90}$ & $55.29$ & $65.53$ & ${59.98}$\\
         \hline
         Reinforce & $70.98$ & $78.81$ & $74.69$ & $58.97$ & $69.05$ & ${63.61}$ & $55.66$ & $63.28$ & ${59.23}$  & ${65.84}$ & $55.31$ & $65.32$ & ${59.90}$ \\
        \bottomrule
        \end{tabular}
    }
    \caption{Results on the CoNLL 2012 test set.}
	% Boldface indicates improvements over the baseline, significant at $p>0.05$, approximate randomization test.}
    \label{tab:conll_data}
\end{table*}
We now investigate the effect of incorporating selectional preferences, implicitly and explicitly, in coreference resolution.  

\begin{figure}[t!]
	\hspace{-1.32em}\includegraphics[width=1.11\columnwidth,keepaspectratio]{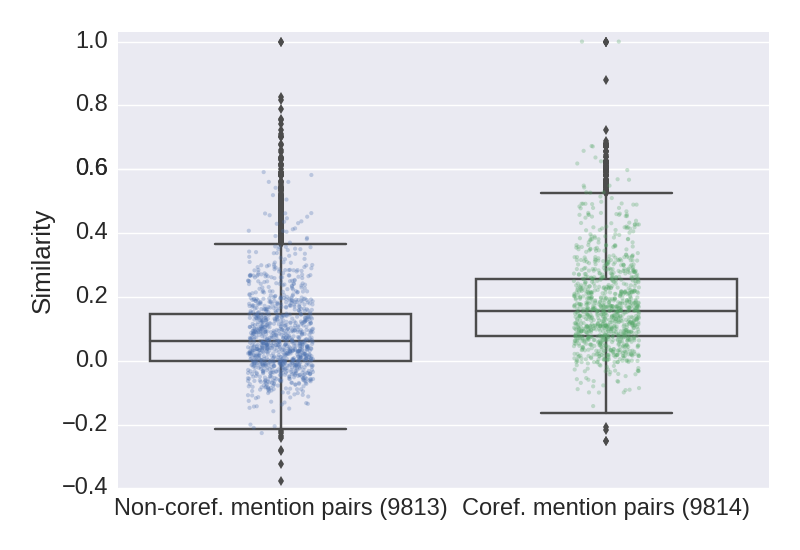}
	\caption{
Selectional preference similarities of 10k coreferent and 10k non-coreferent mention pairs.
%Number of covered pairs in parentheses. 
Lines and boxes represent quartiles, diamonds outliers, points subsamples with jitter.
Coreferent mention pairs are more similar than non-coreferent mention pairs with a Matthews correlation coefficient of $0.30$, indicating weak to moderate correlation.
}
	\label{fig:pair-sim}
\end{figure}

Figure~\ref{fig:pair-sim} shows the selectional preference similarity of $10.000$ coreferent and $10.000$ non-coreferent mention pairs sampled randomly from the CoNLL 2012 training set.
As we can see, while coreferent mention pairs are more similar than non-coreferent mention pairs according to the selectional preference similarity, 
there is not a direct relation between the similarity values and the coreferent relation.
This indicates that coreference does not have a linear relation to the selectional preference similarities.
However, it is worth investigating how these similarity values affect the overall performance 
when they are combined with other knowledge sources in a non-linear way.
%This could explain why selectional preferences do not improve the performance in coreference models with linear feature combinations.

We select the ranking model of deep-coref \cite{clarkkevin16a} as our baseline.
deep-coref is a neural model that combines the input features through several hidden layers. 
\emph{Baseline} in Table~\ref{tab:conll_data} reports our baseline results on the CoNLL 2012 test set.
The results are reported using \emph{MUC} \cite{vilain95}, \emph{B}$^3$ \cite{bagga98b}, \emph{CEAF}$_e$ \cite{luoxiaoqiang05a}, 
the average $F_1$ score of these three metrics, i.e.\ CoNLL score, and \emph{LEA} \cite{moosavi16b}.
\begin{table}[t!]
    \begin{center}\footnotesize
    \resizebox{\columnwidth}{!}{
    \begin{tabular}{l|rrrrr}
     \multicolumn{1}{c}{} & MUC & B$^3$ & CEAF$_e$ & CoNLL & LEA  \\ \hline
     \multicolumn{1}{c}{} & \multicolumn{5}{c}{development} \\ \hline
     baseline & $74.10$ & $63.95$ & $59.73$ & $65.93$ & $60.16$ \\ 
     +embedding & $74.38$ & $64.42$ & $60.45$ & $66.42$ & $60.65$\\ 
     +binned sim. & $74.36$ & $64.54$ & $60.21$ & $66.37$ & $60.77$\\ \hline
     \multicolumn{1}{c}{} & \multicolumn{5}{c}{test} \\ \hline
     baseline &  $74.72$ & $63.26$  &  $58.82$ & $65.60$ & $59.59$ \\ 
     +embedding & $74.53$ & $63.41$ & $59.03$ & $65.66$ & $59.69$ \\ 
     +binned sim. & $74.85$ & ${63.64}$ &${59.21}$ & ${65.90}$ &${59.98}$\\ \hline
    \end{tabular}
    }
    \end{center}
    \caption{Incorporating the selectional preference model as new embeddings (+embedding) vs. as new pairwise features (+binned sim.).}
    \label{tab:embedding_vs_binned}
\end{table}
\begin{table*}[!htbp]
\begin{center}
%\resizebox{\textwidth}{!}{%
\begin{tabular}{l|l}
        %\toprule
        %text & \\ 
        %\midrule
        does [\textbf{that}]$_{ante}$ really impact the case ... [it]$_{ana}$ just shows & {(impact@nsubj,shows@nsubj)} \\ \hline
        %\emph{the protest} revealed an ... \emph{it} was a venting of & \\
        [it]$_{ante}$ will ask a U.S. bankruptcy court to allow [it]$_{ana}$ & {(ask@nsubj,allow@dobj)}\\ \hline
        %[The country \textbf{coroner}]$_{ante}$ says [he]$_{ana}$ urged & (says@nsubj,urged@nsubj) \\ \hline
        [a \textbf{strain} that has n't even presented [itself]$_{ana}$]$_{ante}$ & {(presented@nsubj,presented@dobj)} \\ \hline
        %\bottomrule
\end{tabular}
%    }
    \end{center}
    \caption{Examples of +SP correct links on the development set that do not exist in the baseline output.}
    \label{tab:good_example}
\end{table*} 
deep-coref includes the embeddings of the dependency governor of mentions.
Combined with the relative position of a mention to its governor, 
deep-coref may be able to implicitly capture selectional preferences to some extent.
$-$\emph{gov} in Table~\ref{tab:conll_data} represents deep-coref performance when governors are not incorporated.
As we can see, the exclusion of the governor information does not affect the performance.
%The small differences are not significant. 
This result shows that the implicit modeling of selectional preferences 
does not provide any additional information to the coreference resolver.   

%+\emph{SP} represents the performance of the ranking model in which our selectional preference model is incorporated.
For each mention, we consider (1) the whole mention string, (2) the whole mention string without articles, (3) mention head,
(4) context representation, i.e.\ governor@dependency-relation, and (5) entity types if the mention is a named entity.
We obtain an embedding for each of the above properties if they exist in the selectional preference model, otherwise we set them to unknown.

For each (antecedent, anaphor) pair, we consider all the acquired embeddings of anaphor and antecedent.
We try two different ways of incorporating this knowledge into deep-coref including: 
(1) incorporating the computed embeddings directly as a new set of inputs, i.e. \emph{+embedding} in Table~\ref{tab:embedding_vs_binned}. 
We add a new hidden layer on top of the new embeddings and combine its output with outputs of the hidden layers associated with other sets of inputs;
and (2) computing a similarity value between all possible combinations of the antecedent-anaphor acquired embeddings  
and then binarizing all similarity values, i.e.\ \emph{+binned sim.}\ in Table~\ref{tab:embedding_vs_binned}.

Providing selectional preference embeddings directly to deep-coref adds more complexity to the baseline coreference resolver.
Yet, it performs on-par with \emph{+binned sim.} on the development set
and generalizes worse on the test set.
\emph{+SP} in Table~\ref{tab:conll_data} is the performance of \emph{+binned sim.} on the test set.
As we can see from the results,
adding selectional preferences as binary features improves over the baseline.

\emph{Reinforce} in Table~\ref{tab:conll_data} presents the results of the reward-rescaling model of \newcite{clarkkevin16b} 
that are so far the highest reported results on the official test set.
The reward rescaling model of \newcite{clarkkevin16b} casts the ranking model of \newcite{clarkkevin16a} in the reinforcement learning framework
which considerably increases the training time, from two days to six days in our experiments.

\begin{table}[!]
    \centering
	\small
    %\resizebox{\columnwidth}{!}{
        \begin{tabular}{l|rrr}
			\toprule
			\multirow{2}{*}{Error type} & \multicolumn{3}{c}{Mention type} \\
			& Proper & Common & Pronoun \\
			\midrule
			Recall & -28 & -29 & -53\\
			Precision & +18 & +74 & +61\\
			\bottomrule
        \end{tabular}
    %}
    \caption{Differences in the number of recall and precision errors on the \conll test set in comparison to the baseline.}
    \label{tab:errors}
\end{table}

We analyze how our selectional preference model affects the resolution of various types of mentions.
We use \newcite{martschat14}'s toolkit \footnote{\url{https://github.com/smartschat/cort}} 
to perform recall and error analyses.
The differences in the number of recall and precision errors in +\emph{SP} compared to \emph{baseline}
on the test set are reported in Table~\ref{tab:errors}.

By using our selectional preference features, the number of recall errors decreases for all types of mentions.
The recall error reduction is more prominent for pronouns.
On the other hand, the number of precision errors increases for all types of mentions.
The increase in the precision error is the highest for common nouns.
Overall, +\emph{SP} creates about 260 more links than \emph{baseline}. 

Table~\ref{tab:good_example} lists a few examples from the development set in which +\emph{SP} creates a link that \emph{baseline} does not.
It also includes the similarity that has a high value for the linked mentions and probably is the reason for creating the link. 
For instance, in the first example, based on our model, similarity(\emph{impact@nsubj,shows@nsubj}) is known and 
it is also higher than similarity(\emph{impact@dobj,shows@nsubj}).
%Therefore, \emph{it} is linked to {that}.
%As an example for incorrect links that are created by incorporating selectional preferences, 
%in sentence ``passengers frequently press \emph{the station} for Disney on ticket machines ... to enjoy \emph{the park} when \emph{it} first opens'',
%\emph{it} is linked to \emph{the station} while the correct antecedent is \emph{the park}.
%According to our model \emph{(park,open@nsubj)} and \emph{(station,open@nsubj)} have close similarity values 
%while the similarity of \emph{(enjoy@dobj,open@nsubj)} is lower than that of \emph{(press@dobj,open@nsubj)}.
%Most of the incorrect nominal-nominal links are created between the mentions with the same head or the same governor and dependency relation.

In order to estimate a higher bound on the expected performance boost,
we run the \emph{baseline} and +\emph{SP} models only on anaphoric mentions.
By using anaphoric mentions, the performance
improves by one percent,
based on both the CoNLL score and \emph{LEA}.
This result indicates that the incorporation of selectional preferences creates many links for non-anaphoric mentions, 
which in turn decreases precision.
Therefore, the overall performance does not improve substantially when system mentions are used.
deep-coref incorporates anaphoricity scores at resolution time.
One possible way to further improve the results of +\emph{SP} is to incorporate anaphoricity scores at the input level. 
In this way, the coreference resolver could learn to use selectional preferences 
mainly for mentions that are more likely to be anaphoric. 
However, given that the F$_1$ score of current anaphoricity determiners or singleton detectors is only around $85$ percent \cite{moosavi16a,moosavi17a}, the effect of using system anaphoricity scores might be small.
%We also incorporate our selectional preference model to the ranking model of cort coreference resolver developed by \newcite{martschat15c}.
%cort is a coreference model without any hidden layers.
%The feature combination is manually designed in cort 
%to make the base features useful for the coreference model.
%Since cort does not have any hidden layers to properly combines the selectional preferences with other available information,
%we do not get any improvements from incorporating our model in cort.
%We could only improve the averaged CoNLL score in the gold mention setting, i.e.\ performing coreference resolution on gold mentions instead of all system detected mentions,
%by about 0.4 percent.

\section{Conclusions}
We introduce a new model of selectional preferences, which combines dependency-based word embeddings and fine-grained entity types. 
In order to be effective, a selectional preference model should (1) have a high coverage so it can be used for large datasets like CoNLL,
and (2) be combined with other knowledge sources in a non-linear way. 
Our selectional preference model slightly improves coreference resolution performance, but considering the extra resources that are required to train the model,
it is debatable whether such small improvements are advantageous for solving coreference.

\section*{Acknowledgments} The authors would like to thank the three anonymous reviewers for their helpful comments. 
This work has been supported by the German Research
Foundation (DFG) as part of the Research Training Group
“Adaptive  Preparation  of  Information  from  Heterogeneous  Sources”  (AIPHES) under  grant  No.
GRK 1994/1 and the Klaus Tschira Foundation, Heidelberg,
Germany. 
\bibliographystyle{emnlp_natbib}
\bibliography{add}

\onecolumn
\section*{Appendix}

\begin{figure*}[b!]

\centering
\begin{adjustbox}{max width=\textwidth}
\begin{tabular}{llll}

\toprule
\textbf{Query} & \textbf{Most sim. predicate slots} & \textbf{Most sim. entity types} & \textbf{Most sim. phrases} \\
\midrule
sink@nsubj & sink@nsubj:xsubj & /product/ship & Sea\_Diamond \\
 & sink@nsubjpass & /event/natural\_disaster & Prestige\_oil\_tanker \\
 & sinking@nmod:of & /finance/stock\_exchange & Samina \\
 & slide@nsubj & /astral\_body & Estonia\_ferry \\
 & capsizing@nmod:of & /person/religious\_leader & k-159 \\
 & plunge@nsubj & /finance/currency & Navy\_gunboat \\
 & sink@nmod:along\_with & /military & Dona\_Paz \\
 & sinking@nsubj & /geography/glacier & ferry\_Estonia \\
 & tumble@nsubj & /product/airplane & add-fisk-independent-nytsf \\
 & slip@nsubj & /transit & Al-Salam\_Boccaccio \\

\midrule

ship & capsize@nmod:of & /product/ship & vessel \\
 & some@nmod:aboard & /train & cargo\_ship \\
 & experience@nmod:aboard & /product/airplane & cruise\_ship \\
 & afternoon@nmod:aboard & /transit & boat \\
 & pier@nmod:for & /product/spacecraft & freighter \\
 & escort@nmod:including & /location/bridge & container\_ship\\
 & lift-off@nmod:of & /broadcast/tv\_channel & cargo\_vessel \\
 & disassemble@nsubjpass:xsubj & /location & Navy\_ship \\
 & near-collision@nmod:with & /living\_thing & warship \\
 & Conger@compound & /chemistry & tanker \\

\midrule

steer@dobj & guide@dobj & /broadcast/tv\_channel & business\_way \\
 & steer@nsubjpass & /product/car & newr\_nbkg\_nwer\_ndjn \\
 & shepherd@dobj & /organization/sports\_team & BahrainDinar \\
 & steering@nmod:of & /product/ship & reynard-honda \\
 & nudge@dobj & /product/spacecraft & zigzag\_course \\
 & pilot@dobj & /event/election & team\_home \\
 & propel@dobj & /medicine/medical\_treatment & U.S.\_energy\_policy \\
 & maneuver@dobj & /building/theater & williams-bmw \\
 & divert@dobj & /education/department & interest-rate\_policy \\
 & lurch@nsubj & /product/airplane & trimaran \\

\midrule

/product/ship & Repulse@conj:and & /product/airplane & battleship\_Bismarck \\
 & destroyer@amod & /train & pt\_boat \\
 & capsize@nmod:of & /product/car & battleship \\
 & experience@nmod:aboard & /park & USS\_Nashville \\
 & near-collision@nmod:with	& /military & USS\_Indianapolis \\
 & line@cc & /event/natural\_disaster & k-159 \\
 & brig@conj:and & /award & frigate \\
 & -lrb-@nmod:on & /geography/island & warship \\
 & Umberto@conj:and & /person/soldier & Oriskany \\
 & rumour@xcomp & /location/body\_of\_water & sister\_ship \\

\bottomrule

\end{tabular}
\end{adjustbox}

\caption{Most similar terms for the queries \emph{sink@nsubj}, \emph{ship}, \emph{steer}, and \emph{/product/ship}.}
\label{fig:examples}

\end{figure*}

\end{document}